\documentclass[letterpaper]{article} % DO NOT CHANGE THIS
\usepackage{aaai2026}  % DO NOT CHANGE THIS
\usepackage{times}  % DO NOT CHANGE THIS
\usepackage{helvet}  % DO NOT CHANGE THIS
\usepackage{courier}  % DO NOT CHANGE THIS
\usepackage[hyphens]{url}  % DO NOT CHANGE THIS
\usepackage{graphicx} % DO NOT CHANGE THIS
\usepackage{natbib}  % DO NOT CHANGE THIS AND DO NOT ADD ANY OPTIONS TO IT
\usepackage{caption} % DO NOT CHANGE THIS AND DO NOT ADD ANY OPTIONS TO IT
\usepackage{colortbl} 
\usepackage{xcolor}
\usepackage{array} 
\usepackage{algorithm}
\usepackage{algpseudocode}
\usepackage{booktabs}
\usepackage{multirow}
\definecolor{Color1}{RGB}{210, 170, 230}
\usepackage{amsmath}
\usepackage{amssymb}
\usepackage{breqn}
\usepackage{amsfonts}
\usepackage{newfloat}
\usepackage{listings}
\DeclareCaptionStyle{ruled}{labelfont=normalfont,labelsep=colon,strut=off} % DO NOT CHANGE THIS
\floatstyle{ruled}
\newfloat{listing}{tb}{lst}{}
\floatname{listing}{Listing}
\title{Training-free Token Reduction for Vision Mamba}
\author{
    Qiankun Ma\textsuperscript{\rm 1,\rm 2}, Ziyao Zhang\textsuperscript{\rm 1,\rm 2}, Chi Su\textsuperscript{\rm 3}, 
    Jie Chen\textsuperscript{\rm 1,\rm 3}, Zhen Song\textsuperscript{\rm 1}, Hairong Zheng\textsuperscript{\rm 1,\rm 2,\rm 4}, Wen Gao\textsuperscript{\rm 1,\rm 3}
}
\affiliations{
    \textsuperscript{\rm 1}Peng Cheng Laboratory, \textsuperscript{\rm 2}Shenzhen Institutes of Advanced Technology, Chinese Academy of Sciences \\
    \textsuperscript{\rm 3}Peking University, \textsuperscript{\rm 4}University of the Chinese Academy of Sciences\\
    
    maqiankun2018@gmail.com, zhangzy@pcl.ac.cn 

}

\usepackage{bibentry}
\makeatletter
\gdef\copyright@on{}
\makeatother

\begin{document}

\maketitle

\begin{abstract}
Vision Mamba has emerged as a strong competitor to Vision Transformers (ViTs) due to its ability to efficiently capture long-range dependencies with linear computational complexity. While token reduction, an effective compression technique in ViTs, has rarely been explored in Vision Mamba. Exploring Vision Mamba's efficiency is essential for enabling broader applications. However, we find that directly applying existing token reduction techniques for ViTs to Vision Mamba leads to significant performance degradation. This is primarily because Mamba is a sequence model without attention mechanisms, whereas most token reduction techniques for ViTs rely on attention mechanisms for importance measurement and overlook the order of compressed tokens. In this paper, we investigate a Mamba structure-aware importance score to evaluate token importance in a simple and effective manner. Building on this score, we further propose MTR, a training-free \textbf{M}amba \textbf{T}oken \textbf{R}eduction framework. Without the need for training or additional tuning parameters, our method can be seamlessly integrated as a plug-and-play component across various Mamba models. Extensive experiments demonstrate that our approach significantly reduces computational workload while minimizing performance impact across various tasks and multiple backbones. Notably, MTR reduces FLOPs by approximately 40\% on the Vim-B backbone, with only a 1.6\% drop in ImageNet performance without retraining.
\end{abstract}


\section{Introduction}
In recent years, Transformers have made remarkable progress in the field of computer vision, with Vision Transformers (ViTs)~\cite{dosovitskiy2020image} being a prime example. However, ViTs encounter challenges due to the quadratic growth of self-attention complexity as the input size increases. Mamba~\cite{gu2023mamba}, as a sequence model, has demonstrated substantial potential in addressing these issues, thanks to its linear computational complexity. The emergence of Vision Mamba~\cite{zhu2024vision, liu2024vmambavisualstatespace} has garnered extensive attention and is regarded as a strong competitor to ViTs~\cite{liu2024vision}.
\begin{figure}[!t]
  \centering
  \includegraphics[width=0.48\textwidth]{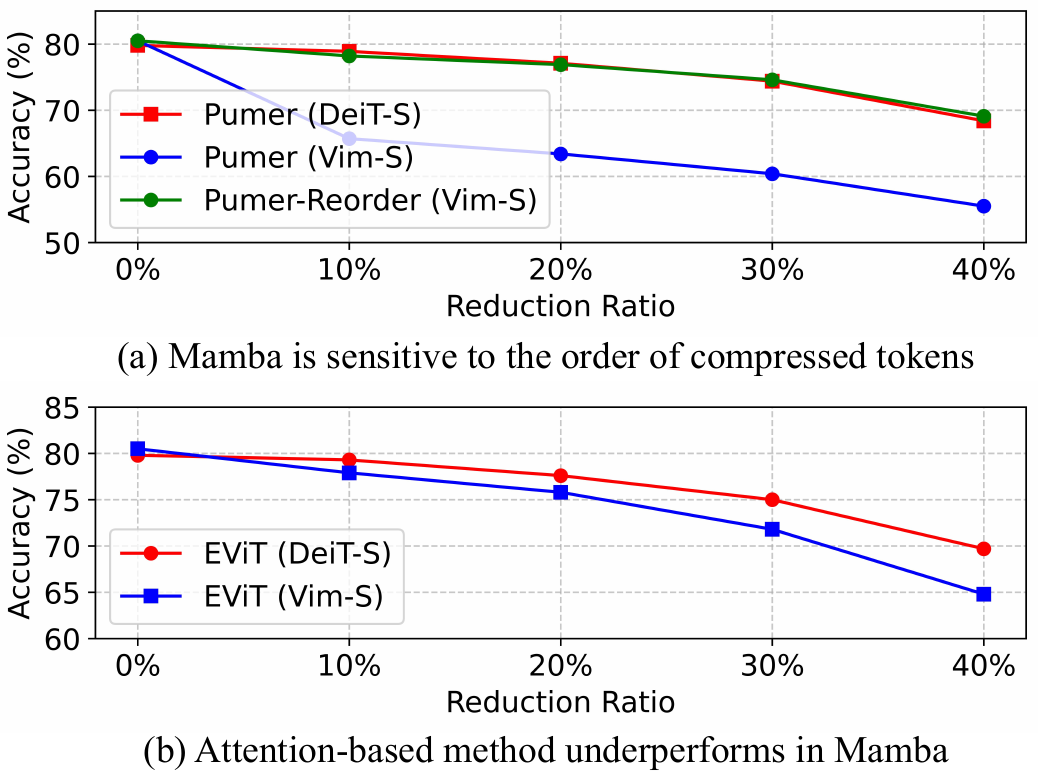}
  % includesvg
  % \includesvg[width=3.0in]{figures/figure1.svg}
  \caption{(a) Mamba is highly sensitive to the sequence of compressed tokens, whereas the transformer is modeled as unordered. The reordering operation can effectively address this issue. (b) Attention-based compression methods (e.g., EViT) tend to underperform in Mamba due to the absence of an attention mechanism in Mamba.}
\label{fig_1}
% \end{wrapfigure}
\end{figure}

Token reduction~\cite{rao2021dynamicvit, pan2021ia, yuan2021tokens, renggli2022learning, chen2023diffrate, meng2022adavit, cao2023pumer, liang2022not} have been shown to be effective in enhancing the efficiency of ViTs, as the token length or number of tokens is independent of the model architecture. 
Consistent with existing research efforts to improve the efficiency of ViTs, exploring the efficiency of Vision Mamba is crucial for enabling real-time applications. Recently, UTR~\cite{zhan2024rethinking} first introduced the training-free token reduction technique to Mamba-based models for natural language processing (NLP) tasks; however, training-free token reduction remains largely unexplored in Vision Mamba.

Given that Vision Mamba processes input tokens by dividing them into patches, similar to ViTs, applying existing ViTs' token reduction techniques to Vision Mamba might seem like a straightforward approach to enhance efficiency. However, as illustrated in Fig.~\ref{fig_1}, directly applying existing token reduction methods for ViTs to Vision Mamba results in significant performance degradation. We attribute this to two main factors. First, Mamba is a sequential model, and the order of compressed tokens significantly impacts performance. As shown in Fig.~\ref{fig_1}(a), using the classical token reduction method Pumer~\cite{cao2023pumer} on Vision Mamba (Vim-S)~\cite{zhu2024vision} and ViT (DeiT-S)~\cite{touvron2021training} backbones on the ImageNet dataset reveals a dramatic performance drop when Pumer is applied directly to Vim-S. This occurs because the compression disrupts the original token order, which can be mitigated by reordering tokens post-compression. 
Second, existing training-free token reduction methods often rely on attention mechanisms and can be categorized into two types: attention-based and CLS token-based. Attention-based methods, such as Zero-TPrune~\cite{wang2024zero} and VoMix~\cite{peng2024vote}, heavily rely on the attention mechanism, frequently utilizing intermediate results from the attention computation, such as the Q, K matrices or the attention maps. Since the Mamba model lacks these intermediate results, such methods cannot be directly migrated to the Mamba model. Another type, CLS token-based methods, such as EViT~\cite{liang2022not}, uses the [CLS] token's attention score to measure token importance, a highly effective approach in ViT token compression~\cite{wang2024cls,haurum2023tokens, zhang2024cls}, and can be directly applied to Mamba. However, since Mamba lacks an attention mechanism, we substitute attention scores with token similarity. Consequently, when applying EViT to Vision Mamba, we use the similarity between the [CLS] token and other tokens to assess importance. As depicted in Fig.~\ref{fig_1}(b), Vision Mamba's performance is notably inferior to ViT's, especially at high reduction rates (e.g., a 40\% compression rate results in a 4.9\% performance gap), highlighting a substantial discrepancy.

This observation prompted us to consider whether Mamba possesses an “attention score”, a indicator that assesses token importance without incurring additional computational overhead. Through extensive analysis and experimentation, we found that the timescale parameter $\Delta$ in Mamba effectively serves this purpose. Building on this insight, we developed a training-free Mamba token reduction framework, named MTR. Specifically, MTR first evaluates each token’s importance using timescale parameter $\Delta$ and groups them according to importance level. We then merge the least important tokens with those in a specific grouping based on similarity to accomplish the compression process.
Our approach is generalizable across tasks and applicable to any Mamba-based model. To the best of our knowledge, we are the first to explore Mamba structure-aware token evaluation scores and to propose a training-free Mamba token reduction framework.
Empirically, our method can significantly reduce computational demands while maintaining competitive accuracy without any retraining. We summarize our contributions as follows:
\begin{itemize}
\itemsep=-1pt
    \item We identified that directly applying existing token reduction techniques from ViTs to Vision Mamba leads to significant performance degradation. Analysis revealed two primary reasons. First, Mamba is a sequential model, and token order significantly impacts performance, which can be mitigated by reordering tokens. Second, existing token reduction methods rely on attention mechanisms, which Mamba lacks. To address this, we explored Mamba's internal "attention score" and found that the timescale parameter $\Delta$ can effectively assess token importance.

    \item Based on our exploration, we developed a training-free Mamba token reduction framework MTR. MTR 
    first evaluates token importance using Mamba structure-aware scores, followed by asymmetric grouping based on the computed importance. Finally, it merges the least important tokens with those in a specific grouping to achieve token reduction.

    \item Extensive experiments show that MTR significantly reduces computational workload while maintaining competitive accuracy across various tasks and multiple backbones. For instance, on the Vim-B backbone, it reduces FLOPs by ~40\% with only a 1.6\% drop in ImageNet performance, without retraining.
\end{itemize}

\section{Related Work}
\label{sec:related_work}
\subsection{Vision Mamba}
Mamba~\cite{gu2023mamba}, an extension of state space model (SSM)~\cite{gu2021efficiently,smith2022simplified,mehta2022long, fu2022hungry, wang2023selective}, has achieved excellent performance in NLP tasks. Its ability to capture long-range dependencies with linear computational complexity has led many researchers to adapt it for visual tasks~\cite{chen2024rsmamba, guo2024mambair, hatamizadeh2024mambavision, li2024videomamba, patro2024simba, pei2024efficientvmamba, qiao2024vl, ruan2024vm, shi2024multi, yang2024vivim, zhan2024exploring, behrouz2024mambamixer}. For example, ViM~\cite{zhu2024vision} incorporates a bidirectional SSM module and constructs an isotropic architecture similar to ViT~\cite{dosovitskiy2020image}. VMamba~\cite{liu2024vmambavisualstatespace} introduces a cross-scan module, creating a hierarchical SSM-based architecture. PlainMamba~\cite{yang2024plainmamba} enhances spatial continuity through continuous 2D scanning, ensuring token adjacency in the scanning sequence. LocalMamba~\cite{huang2024localmamba} uses a local scanning strategy to capture local dependencies. However, most of these studies focus on Mamba's structure and scanning mechanisms, with limited exploration of model inference efficiency. Our proposal effectively accelerates Vision Mamba’s inference through token reduction, offering a simple, training-free, and plug-and-play solution for various Mamba-based models.

\subsection{Token Reduction}
Token reduction is a highly effective strategy to enhance computational efficiency by reducing the number of processed tokens or patches. It has shown significant potential in accelerating Transformers in both natural language processing~\cite{goyal2020power, kim2020length, kim2022learned} and computer vision~\cite{fayyaz2022adaptive, meng2022adavit, rao2021dynamicvit, song2022cp, yin2022vit, bolya2022token, kong2022spvit, dou2023tore, marin2021token, ryoo2021tokenlearner, xu2022groupvit, shang2024llava, shen2025numerical, xu2025rethinking}. For example, EViT~\cite{liang2022not} identifies informative tokens based on the [CLS] token, thereby simplifying the training process. PuMer~\cite{cao2023pumer} introduced a token reduction framework for large-scale vision-language models (VLMs) that employs text-informed pruning and modality-aware merging strategies to progressively reduce the number of input image and text tokens. ToMe~\cite{bolya2022token} determines token redundancy by measuring the dot product similarity between token keys and merges tokens accordingly.

However, token reduction techniques remain largely unexplored in Mamba. As a sequence model lacking the attention mechanism found in transformers, Mamba is not directly compatible with existing transformer-based token reduction methods. To our knowledge, HSA~\cite{zhan2024exploring} was the first to investigate token compression in Vision Mamba, achieving this through importance-based token cropping and retraining. Nonetheless, the exploration of an “attention score” in Mamba and the development of a training-free approach remain uncharted territories. Our method not only clarifies why prior token reduction techniques are unsuitable for Mamba but also thoroughly examines the “attention score” for assessing token importance within the Mamba framework. Furthermore, we introduce a simple, effective, and training-free solution that both accelerates and restores the performance of compressed Mamba models.

\section{Methodology}
\subsection{Preliminary}
% \textbf{State space model (SSM).}
The classical state space model (SSM) is a continuous system that employs an implicit hidden state $h(t)\in\mathbb{R}^{N\times 1}$ to transform a 1-D sequence input $x(t)\in\mathbb{R}$ into an output $y(t)\in\mathbb{R}$, which can be written as follows:
\begin{equation}\label{eq:cts_ssm}
\begin{split}
    &{h}'(t)=\mathbf{A}h(t)+\mathbf{B}x(t), \\
    &y(t)=\mathbf{C}h(t)+\mathbf{D}x(t),
\end{split}
\end{equation}
where $\mathbf{A}\in {\mathbf{R}^{N\times N}}$ denotes the evolution matrix, while $\mathbf{B}\in {\mathbf{R}^{N\times 1}}$ and $\mathbf{C}\in {\mathbf{R}^{1\times N}}$ serve as the projection parameters, and the skip connection $\mathbf{D}\in\mathbb{R}$. 

SSM faces great challenges when integrated into deep learning algorithms due to its continuous-time nature. To be effectively applied to deep neural networks, SSM must first be transformed into their discrete counterparts through zero-order hold (ZOH) discretization. Specifically, the continuous parameters $\mathbf{A}, \mathbf{B}$ are converted into their discretized versions $\overline{\mathbf{A}}, \overline{\mathbf{B}}$ using a timescale parameter $\Delta\in\mathbb{R}$:
\begin{equation}\label{eq:AB_discretization}
    \begin{split}
    &\overline{\mathbf{A}} = \mathrm{exp}(\Delta \mathbf{A}),\\
    &\overline{\mathbf{B}} = (\Delta \mathbf{A})^{-1}(\mathrm{exp}(\Delta \mathbf{A})-\mathbf{I})\cdot\Delta \mathbf{B}.
    \end{split}
\end{equation}
After obtaining the discretized $\overline{\mathbf{A}}$ and $\overline{\mathbf{B}}$, the discrete SSM rewrite Eq.~\ref{eq:cts_ssm} as follows:
\begin{equation}\label{eq:cts_ssm_dis}
\begin{split}
    &h_t=\overline{\mathbf{A}}h_{t-1}+\overline{\mathbf{B}}x_t, \\
    &y_t=\mathbf{C}h_t+\mathbf{D}x_t.
\end{split}
\end{equation}
Mamba~\cite{gu2023mamba} enhances the SSM by introducing selection, thereby proposing the selective state space model. In this model, the parameters $\mathbf{B}, \mathbf{C}, \Delta$ are directly derived from the input data $x_t$, making them input-dependent parameters $\mathbf{B}_t, \mathbf{C}_t, \Delta_t$. Consequently, the discretized parameters $\overline{\mathbf{A}}_t=\mathrm{exp}(\Delta_t \mathbf{A}),\ \overline{\mathbf{B}}_t=\Delta_t \mathbf{B}_t$ are also input-dependent. The selective state space model is formulated as:
\begin{equation}\label{eq:cts_ssm_select}
\begin{split}
    &h_t=\overline{\mathbf{A}}_{t}h_{t-1}+\overline{\mathbf{B}}_{t}x_t, \\
    &y_t=\mathbf{C}_{t}h_t+\mathbf{D}x_t.
\end{split}
\end{equation}
Mamba practically sets $\mathbf{A}, \overline{\mathbf{A}}_t$ as diagonal matrices. Therefore, $\overline{\mathbf{A}}_th_{t-1}\!=\widetilde{\mathbf{A}}_t\odot h_{t-1}$, where $\odot$ denotes the Hadamard product, and $\widetilde{\mathbf{A}}_t=\mathrm{diag}(\overline{\mathbf{A}}_t)\in\mathbb{R}^{N\times 1}$ represents the matrix composed of the diagonal elements of $\overline{\mathbf{A}}_t$. Additionally, given $\overline{\mathbf{B}}_t=\Delta_t \mathbf{B}_t$ with $\Delta_t\in \mathbb{R}$, we have: 
\begin{equation}\label{eq:cts_ssm_B}
\begin{split}
    &\overline{\mathbf{B}}_{t}x_t=\Delta_t \mathbf{B}_t x_t=\mathbf{B}_t (\Delta_t\odot x_t).
\end{split}
\end{equation}
Similarly, $\mathbf{D} x_t=\mathbf{D}\odot x_t$. Consequently, we can rewrite Eq.~\ref{eq:cts_ssm_select} as:
\begin{equation}\label{eq:modified_ssm_select}
\begin{split}
    &h_t=\widetilde{\mathbf{A}}_{t} \odot h_{t-1}+\mathbf{B}_t (\Delta_t\odot x_t), \\
    &y_t=\mathbf{C}_{t}h_t+\mathbf{D}\odot x_t,
\end{split}
\end{equation}
where $\mathbf{B}_t, \mathbf{C}_t, \Delta_t$ are all derived from the input. Specifically, Mamba uses the following formulas to generate these parameters: $\mathbf{B}_t=(xW_B)^\top,\  \mathbf{C}_t=xW_C,\ \Delta_t=\mathrm{Softplus}(xW_1W_2)$, where $W_B$, $W_C$, $W_1$ and $W_2$ serve as projection matrices.

\subsection{Assessing Token Importance in Vision Mamba}
As previously stated, we aim to explore the “attention score” in Mamba, which measures token importance without requiring additional computation. Given that $\mathbf{B}_t=(xW_B)^\top,  \mathbf{C}_t=xW_C,\ \Delta_t=\mathrm{Softplus}(xW_1W_2)$, $\mathbf{B}_t, \mathbf{C}_t, \Delta_t$, all these parameters are derived from the input $x$ and can serve as token importance assessment scores without incurring extra computational costs. Further analysis reveals that $\Delta_t$ can be viewed as an input gate that modulates the weight of the current input token $x_t$~\cite{han2024demystify}. Specifically, a larger $\Delta_t$ indicates greater focus on the current input, whereas a smaller $\Delta_t$ suggests more reliance on historical memory. The properties of $\Delta_t$ align well with the desired characteristics of an importance score, and thus, we select $\Delta_t$ to evaluate token importance in this study. Additionally, we demonstrate the superiority of $\Delta_t$ over other indicators ($\mathbf{B}_t$, $\mathbf{C}_t$, etc.) through ablation experiments in Section~\ref{exp_4}.

\begin{figure*}[t]
    \centering
    \includegraphics[width=0.95\linewidth]{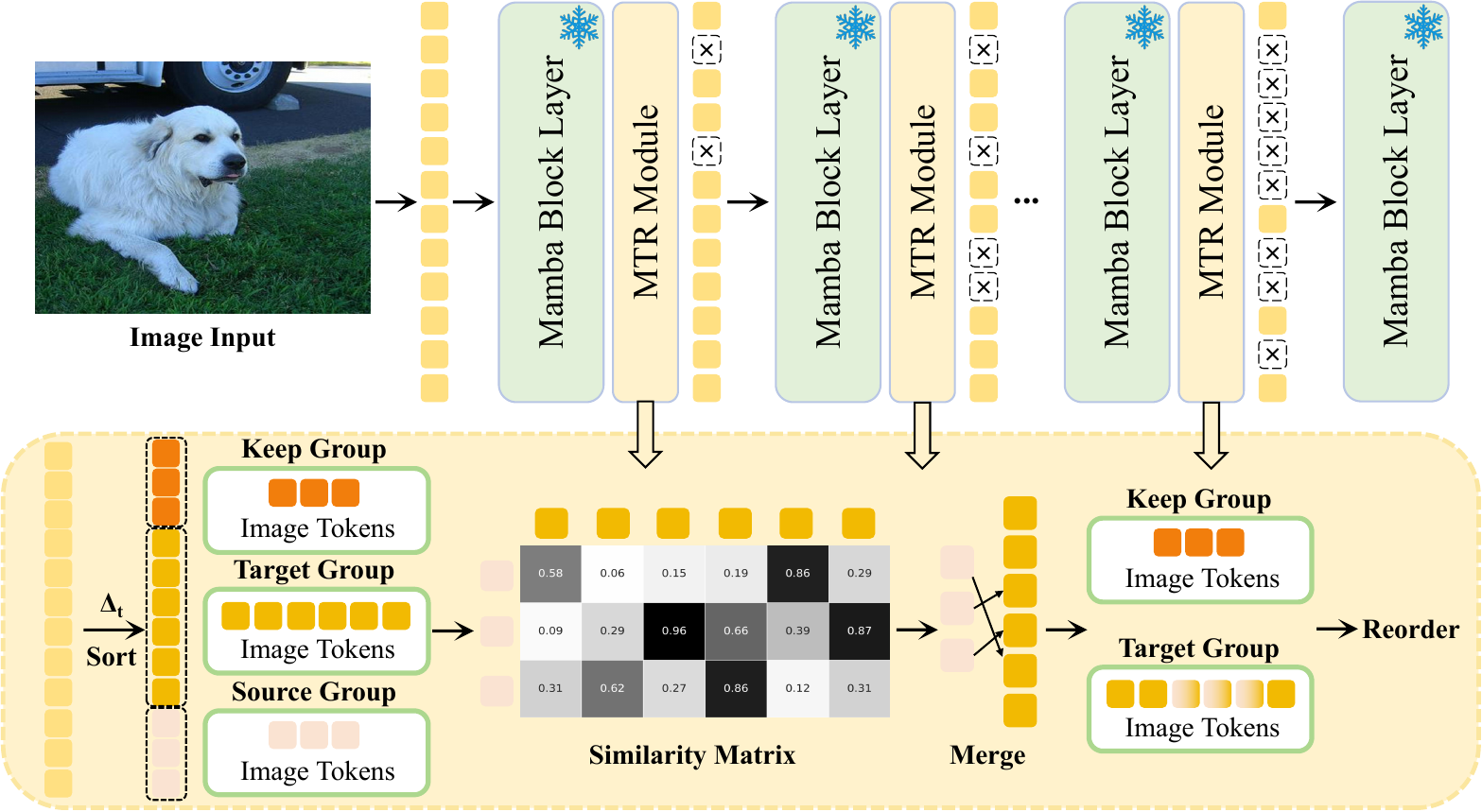}
    \caption{Overview of our proposed framework MTR. Image tokens are processed by the Mamba block and subsequently sorted in descending order according to their importance scores. The tokens are divided into three categories: 'Keep', 'Target', and 'Source'. 'Source' tokens are merged with the most similar 'Target' tokens based on feature similarity. Finally, the remaining tokens are sorted by their original index order and passed to the next Mamba block for further processing.
    }
    \label{fig_2}
\end{figure*}

For the $l^{th}$ layer of the Vision Mamba model, the input token sequence $x^{l} \in \mathbb{R}^{B \times L \times D}$ is transformed into the output $y^{l} \in \mathbb{R}^{B \times L \times D}$ using the following formulation:
\begin{equation} \label{eq:layer_output}
% \vspace{-2mm}
    y^{l} = Linear^{T}( \sum_{s \in \mathbf{S}}SSM_s(x^l)),
\end{equation}
where $S$ denotes the set of scanning heads. For simplicity, residual connections are omitted here.  Each scanning head represents a distinct SSM module with a specific scanning pattern; for instance, ViM~\cite{zhu2024vision} employs two scanning heads: forward and backward. To quantify the importance of each token, we first aggregate $\Delta_t$ across scanning heads, resulting in $\Delta^l =  \sum_{s \in \mathbf{S}}{\Delta_t}_s$. Given that SSM leverages its extensive channel capacity to enable a more nuanced attention distribution, thereby enhancing the model's ability to discern subtle features and interactions among tokens~\cite{zhan2024exploring}, we compute the average of $\Delta^l$ across the last dimension (i.e., the feature dimension $D$) to evaluate token importance:
\begin{equation} \label{eq:importance_metric}
% \vspace{-2mm}
    s^{l} =  \frac{\sum_{d=1}^{D}{\Delta^l_d}}{D}.
\end{equation}
We employ $s^{l} \in \mathbb{R}^{B \times L \times 1}$ as the token importance indicator corresponding to $B\times L$ tokens to guide the reduction process.

\subsection{Importance-Based Token Grouping and Compression}
Based on the importance scores calculated in Eq.~\ref{eq:importance_metric}, we evaluate the significance of different tokens to enable effective token reduction. Accordingly, we have designed a training-free framework, MTR, as illustrated in Fig.~\ref{fig_2}.
We progressively compress the number of tokens by integrating the MTR module after the Mamba block. Specifically, each MTR module first uses the importance scores from Eq.~\ref{eq:importance_metric} to assess each token and then classifies them into three groups: 'keep,' 'target,' and 'source,' ordered by importance from highest to lowest. Tokens in the 'source' group are merged with those in the 'target' group, while the critical tokens in the 'keep' group remain unchanged. Through this asymmetric grouping, we both protect the core knowledge from alteration and reduce computational costs in similarity calculations. Fig.~\ref{fig_3} vividly demonstrates the superiority of our grouping method over other methods.
For simplicity, we set the token ratios for the 'keep' and 'source' groups to $k\%$, and the 'target' group to $1-2k\%$. Here, $k$ is a reduction parameter determined by the desired compression ratio. Subsequently, MTR merges the 'source' tokens with the most similar tokens in the 'target' group. Finally, the remaining tokens are reordered according to their original sequential positions. The token reduction algorithm is described in Algorithm~\ref{algo:tip}.

\begin{algorithm}[h]
\caption{Token Reduction Process}
\label{algo:tip}
\textbf{Input:} token sequence $y^{l-1}$, importance scores $s^{l-1}$, token reduction ratio $k$. \\
\textbf{Output:} token sequence $x^{l}$
\begin{algorithmic}[1]
\State Sort the token sequence $y^{l-1}$ in descending order based on importance scores $s^{l-1}$;
\State Calculate the number of tokens in the 'target' group $k^{\prime}$, $k^{\prime}=(1-2k)|y^{l-1}|$;
\State Divide the tokens into three groups: 'keep' $\mathbf{K}$, 'target' $\mathbf{T}$ and 'source' $\mathbf{S}$.
\State Merge 'source' tokens $\mathbf{S}$ into 'target' tokens $\mathbf{T}$ using bipartite soft matching: $\mathbf{T}=\text{bipartite\_merge}(\mathbf{S}, \mathbf{T})$;
\State Aggregate and reorder the remaining tokens: $x^{l}=reorder(concat(\mathbf{K},\mathbf{T}))$  
\Procedure{bipartite\_merge}{ $\mathbf{S}$, $\mathbf{T}$}
    \State For each token $\mathbf{S}_a$ in $\mathbf{S}$, compute its top-1 similar \newline
    \hspace*{\algorithmicindent}token $\mathbf{T}_b$ in $\mathbf{T}$, save the indices $a$ and $b$ into a token \newline\hspace*{\algorithmicindent}edge (an edge between $\mathbf{S}_a$ and $\mathbf{T}_b$), store all token 
    \newline\hspace*{\algorithmicindent}edges in a set $\mathbf{P}$
    \State For each token edge ($a$, $b$) in $\mathbf{P}$, collect tokens from\newline\hspace*{\algorithmicindent}$\mathbf{S}$ and $\mathbf{T}$ connected by the edge, merge these tokens \newline\hspace*{\algorithmicindent}by computing the mean of their token vectors
    \State output: merged tokens $\mathbf{T}$
\EndProcedure
\end{algorithmic}
\end{algorithm}

Notably, according to the above design, token reduction can be performed at any layer, and the grouping ratio $k$ is fixed across all layers. $k$ is not a hyperparameter that requires manual adjustment but is determined by the desired compression ratio.

\section{Experiments}
\subsection{Implementation Details}
We conducted comprehensive experiments on the ImageNet-1K~\cite{deng2009imagenet} classification task, reporting top-1 accuracy (\%). ViM~\cite{zhu2024vision} and VideoMamba~\cite{li2024videomamba} are used as baseline Mamba models. Given that our method is training-free, we adopted the inference techniques from prior work~\cite{zhan2024rethinking} and applied varying FLOPS reduction ratios to the models to validate our approach's effectiveness. All experiments are performed on four NVIDIA V100 GPUs.

\begin{table}[t!]
\small
\centering
\begin{tabular}{l|c|c|c|c}
\toprule
\multirow{2}{*}{Method} & FLOPS   & \multirow{2}{*}{Params (M)}     & Top-1 & \multirow{2}{*}{$\Delta$}
\\
& Reduction & & Acc. (\%) & \\
\toprule
ViM-S              &  0\% & 26 & 80.5   & 0.0    \\
\midrule
+ EViT           & \multirow{5}{*}{20\%}     & 26        & 75.8 & 4.7$\downarrow$   \\
+ PuMer               &       & 26       &76.9 & 3.6$\downarrow$         \\
+ UTR               &       & 26       &77.3 & 3.2$\downarrow$         \\
+ HSA               &       & 26       &76.7 & 3.8$\downarrow$         \\
\rowcolor{blue!12}+ \textbf{MTR}               &       & \textbf{26}     & \textbf{78.8} & \textbf{1.7$\downarrow$}      \\
\midrule
+ EViT           & \multirow{5}{*}{30\%}     & 26        & 71.8 & 8.7$\downarrow$  \\
+ PuMer               &      & 26        & 74.6 & 5.9$\downarrow$      \\
+ UTR               &       & 26       &75.0 & 5.5$\downarrow$         \\
+ HSA               &       & 26       &74.8 & 5.7$\downarrow$         \\
\rowcolor{blue!12}+ \textbf{MTR}               &   & \textbf{26}         &\textbf{77.7} &\textbf{2.8$\downarrow$}     \\
\midrule
+ EViT           & \multirow{5}{*}{40\%}   & 26          &64.8 &15.7$\downarrow$  \\
+ PuMer               &      & 26        &69.1 &11.4$\downarrow$      \\
+ UTR               &       & 26       &71.5 & 9.0$\downarrow$         \\
+ HSA               &       & 26       &71.2 & 9.3$\downarrow$         \\
\rowcolor{blue!12}+ \textbf{MTR}               &   & \textbf{26}         &\textbf{75.4} &\textbf{5.1$\downarrow$}     \\
\toprule
ViM-B              & 0\% & 98 & 81.9   & 0.0    \\
\midrule
+ EViT           & \multirow{5}{*}{20\%}      & 98      & 80.4 &1.5$\downarrow$  \\
+ PuMer               &        & 98      &79.9 &2.0$\downarrow$  \\
+ UTR               &       & 98       &80.4 & 1.5$\downarrow$         \\
+ HSA               &       & 98       &80.1 & 1.8$\downarrow$         \\
\rowcolor{blue!12}+ \textbf{MTR}               &   & \textbf{98}         &\textbf{81.2} &\textbf{0.7$\downarrow$}     \\
\midrule
+ EViT           & \multirow{5}{*}{30\%}   & 98          &78.9 &3.0$\downarrow$  \\
+ PuMer               &     & 98        &78.9 &3.0$\downarrow$         \\
+ UTR               &       & 98       &79.2 & 2.7$\downarrow$         \\
+ HSA               &       & 98       &79.1 & 2.8$\downarrow$         \\
\rowcolor{blue!12}+ \textbf{MTR}               &   & \textbf{98}         &\textbf{81.0} &\textbf{0.9$\downarrow$}     \\

\midrule

+ EViT           & \multirow{5}{*}{40\%}   & 98          &75.9 &6.0$\downarrow$   \\
+ PuMer               &   & 98           &76.8 &5.1$\downarrow$         \\
+ UTR               &       & 98       &78.0 & 3.9$\downarrow$         \\
+ HSA               &       & 98       &77.7 & 4.2$\downarrow$         \\
\rowcolor{blue!12}+ \textbf{MTR}               &   & \textbf{98}         &\textbf{80.3} &\textbf{1.6$\downarrow$}     \\
\bottomrule
\end{tabular}
\caption{Main results of the training-free performance on ViM-S and ViM-B. We compared our method with baseline token reduction methods and evaluated them on the ImageNet-1K dataset under 20\%, 30\%, and 40\% FLOPS reduction.}
\label{tab:main_results_vim}
\end{table}
\subsection{Comparison Methods}
    Following previous studies~\cite{zhan2024rethinking, zhan2024exploring}, we compare our method with PuMer~\cite{cao2023pumer} and EViT~\cite{liang2022not}, two representative transformer token reduction methods. 
    To date, no training-free Vision Mamba token reduction methods have been developed. Consequently, we compare with two existing Mamba token reduction methods: UTR~\cite{zhan2024rethinking} and HSA~\cite{zhan2024exploring}. UTR is designed for NLP tasks, while HSA is a Mamba token pruning method that involves retraining. To ensure fair comparisons, we evaluate these methods in a training-free setting. \textit{Notably, we also compare our method with state-of-the-art token reduction methods in ViT and include comparisons on other tasks in the Appendix.}

\begin{table}[t!]
\small
\centering
\begin{tabular}{l|c|c|c|c}
\toprule
\multirow{2}{*}{Method} & FLOPS   & \multirow{2}{*}{Params (M)}     & Top-1 & \multirow{2}{*}{$\Delta$}
\\ 
& Reduction & & Acc. (\%) & \\
\toprule
VideoM-S              &  0\% & 26 & 81.2   & 0.0    \\
\midrule
+ EViT           & \multirow{5}{*}{20\%}     & 26        & 78.2 & 2.8$\downarrow$   \\
+ PuMer               &       & 26       &78.4 & 3.0$\downarrow$         \\
+ UTR               &       & 26       &78.9 & 2.3$\downarrow$         \\
+ HSA               &       & 26       &79.0 & 2.2$\downarrow$         \\
\rowcolor{blue!12}+ \textbf{MTR}               &       & \textbf{26}     & \textbf{80.2} & \textbf{1.0$\downarrow$}      \\
\midrule
+ EViT           & \multirow{5}{*}{30\%}     & 26        & 75.5 & 5.7$\downarrow$  \\
+ PuMer               &      & 26        & 76.2 & 5.0$\downarrow$      \\
+ UTR               &       & 26       &77.1 & 4.1$\downarrow$         \\
+ HSA               &       & 26       &77.2 & 4.0$\downarrow$         \\
\rowcolor{blue!12}+ \textbf{MTR}               &   & \textbf{26}         &\textbf{79.0} &\textbf{2.2$\downarrow$}     \\
\midrule
+ EViT           & \multirow{5}{*}{40\%}   & 26          &70.2 &11.0$\downarrow$  \\
+ PuMer               &      & 26        &71.1 &10.1$\downarrow$      \\
+ UTR               &       & 26       &74.1 & 7.1$\downarrow$         \\
+ HSA               &       & 26       &74.0 & 7.2$\downarrow$         \\
\rowcolor{blue!12}+ \textbf{MTR}               &   & \textbf{26}         &\textbf{76.6} &\textbf{4.6$\downarrow$}     \\
\toprule
VideoM-B             & 0\% & 98 & 82.7   & 0.0     \\
\midrule
+ EViT           & \multirow{5}{*}{20\%}      & 98      & 80.4 &2.3$\downarrow$  \\
+ PuMer               &        & 98      &81.8 &0.9$\downarrow$  \\
+ UTR               &       & 98       &82.0 & 0.7$\downarrow$         \\
+ HSA               &       & 98       &82.0 & 0.7$\downarrow$         \\
\rowcolor{blue!12}+ \textbf{MTR}               &   & \textbf{98}         &\textbf{82.4} &\textbf{0.3$\downarrow$}     \\
\midrule
+ EViT           & \multirow{5}{*}{30\%}   & 98          &77.7 &5.0$\downarrow$  \\
+ PuMer               &     & 98        &80.5 &2.2$\downarrow$         \\
+ UTR               &       & 98       &81.0 & 1.7$\downarrow$         \\
+ HSA               &       & 98       &81.2 & 1.5$\downarrow$         \\
\rowcolor{blue!12}+ \textbf{MTR}               &   & \textbf{98}         &\textbf{81.7} &\textbf{1.0$\downarrow$}     \\

\midrule

+ EViT           & \multirow{5}{*}{40\%}   & 98          &73.7 &9.0$\downarrow$   \\
+ PuMer               &   & 98           &78.4 &4.3$\downarrow$         \\
+ UTR               &       & 98       &79.4 & 3.3$\downarrow$         \\
+ HSA               &       & 98       &79.6 & 3.1$\downarrow$         \\
\rowcolor{blue!12}+ \textbf{MTR}               &   & \textbf{98}         &\textbf{80.5} &\textbf{2.2$\downarrow$}     \\
\bottomrule
\end{tabular}
\caption{Main results of the training-free performance on VideoMamba-S and VideoMamba-B. We compared our method with baseline token reduction methods and evaluated them on the ImageNet-1K dataset under 20\%, 30\%, and 40\% FLOPS reduction.}
\label{tab:main_results_videom}
\end{table}
\subsection{Main Results}
\textbf{Evaluation on ViM.} As shown in Table~\ref{tab:main_results_vim}, we compare the performance of MTR with baseline methods on the ViM backbone. To ensure a fair comparison, all methods perform token reduction followed by token reordering. It is evident that MTR consistently outperforms all baselines under the same FLOPS reduction ratios. Notably, with a 40\% FLOPS reduction on the ViM-S backbone, MTR outperforms UTR and HSA by 3.9\% and 4.2\%, respectively. For the more robust ViM-B backbone, the performance drop due to token reduction is relatively smaller. Even so, our approach still has a significant advantage over other methods. For instance, at a 40\% FLOPS reduction ratio, our method only decreases by 1.6\%, while UTR and HSA decrease by 3.9\% and 4.2\%, respectively.

\begin{table}[t]
\small
\centering
    \begin{tabular}{c|c|c|c|c}
        \toprule
        \multirow{2}{*}{Model} & \multirow{2}{*}{Indicator}   & 20\%    & 30\%  & 40\% 
        \\
        &    &  Reduction  &  Reduction  & Reduction
        \\
        \toprule
        \multirow{5}{*}{ViM-S} 
         &
        [CLS] 
        &
        77.0 
        &
        74.4 
        &
        68.8
        \\
        &
        $\mathbf{X}_{t}$ 
        &
        77.9 
        &
        76.1 
        &
        72.6
        \\
        &
        $\mathbf{C}_{t}$ %
        &
        78.1
        &
        76.5
        &
        73.3
        \\
        &
        $\textbf{B}_{t}$ %
        &
        78.1 
        &
        77.1 
        &
        74.5
        \\
        \rowcolor{blue!12}&
        \textbf{$\mathbf{\Delta}_{t}$} %
        &
        \textbf{78.8}
        &
        \textbf{77.7}
        &
        \textbf{75.4}
        \\
            \midrule 
        \multirow{5}{*}{ViM-B} %
        &
        [CLS]
        &
        80.9 %
        &
        79.8 %
        &
        77.8
        \\
        &
        
        $\mathbf{X}_{t}$ %
        &
        80.5
        &
        80.3
        &
        79.5
        \\
        &
        $\mathbf{C}_{t}$ %
        &
        80.7
        &
        80.0
        &
        78.8
        \\
        &
        $\textbf{B}_{t}$ %
        &
        80.7
        &
        80.1
        &
        79.2
        \\
        \rowcolor{blue!12}&
        \textbf{$\mathbf{\Delta}_{t}$} %
        &
        \textbf{81.2}
        &
        \textbf{81.0}
        &
        \textbf{80.3}
        \\
    \bottomrule
    \end{tabular}
    \caption{
    Ablation study on the impact of different indicator choices on top-1 accuracy (\%). $\mathbf{X}_{t}$ means using hidden state features as importance indicator. [CLS] indicates that we use the similarity between the [CLS] token and other tokens as an importance assessment. The timescale parameter $\mathbf{\Delta}_{t}$ offers a better measure of token importance than other indicators.
    }
    \label{tab:exp_ab_indicator}
    \vspace{-0.1cm}
\end{table}

\noindent \textbf{Evaluation on VideoMamba.} In Table~\ref{tab:main_results_videom}, we present the performance of our method compared to baseline methods on the VideoMamba backbone. Consistent with previous findings, MTR outperforms all baselines, further demonstrating the effectiveness of our approach. Notably, the EViT method underperforms compared to other methods in most cases, primarily because it relies on the attention mechanism for importance measurement, which, as discussed earlier, leads to significant performance degradation when applied to Mamba.

\subsection{Ablation Studies.}
\label{exp_4}
\textbf{Analysis on importance indicator.} To comprehensively evaluate the most effective token importance measures in Mamba, we explored various importance indicators, as shown in Table~\ref{tab:exp_ab_indicator}. Clearly, $\mathbf{\Delta}_{t}$ outperforms other indicators when used as an importance indicator, which aligns with our previous analysis. Additionally, using $\textbf{B}_{t}$ as an importance indicator intuitively yields good performance; as in Eq.~\ref{eq:modified_ssm_select}, both $\textbf{B}_{t}$ and $\mathbf{\Delta}_{t}$ directly influence the sequence input $x_t$, providing a better measure of token importance. We believe that jointly considering $\textbf{B}_{t}$ and $\mathbf{\Delta}_{t}$ could offer an even better measure of token importance in Mamba, and we leave this exploration for future work.
Furthermore, the [CLS] token, an effective indicator of token importance in transformers~\cite{wang2024cls,haurum2023tokens, zhang2024cls}, underperforms in the Mamba model. We speculate this is because Mamba is a sequential model, and token positions affect token similarity. For instance, tokens neighboring the [CLS] token naturally exhibit higher similarity, which is unlike in transformers.

\begin{table}[t]
\small
\centering
    \begin{tabular}{c|c|c|c|c}
        \toprule
        \multirow{2}{*}{Model} & \multirow{2}{*}{Strategy}   & 20\%    & 30\%  & 40\% 
        \\
        &    &  Reduction  &  Reduction  & Reduction
        \\
        \toprule
        \multirow{3}{*}{ViM-S} 
         &
        Pruning
        &
        78.5
        &
        77.1 
        &
        74.0
        \\
        &
        Hybrid
        &
        78.6 
        &
        77.5 
        &
        74.6
        \\
        \rowcolor{blue!12}&
        \textbf{Merging} %
        &
        \textbf{78.8}
        &
        \textbf{77.7}
        &
        \textbf{75.4}
        \\
            \midrule 
        \multirow{3}{*}{ViM-B} %
        &
        Pruning
        &
        81.1 
        &
        80.8 
        &
        80.1
        \\
        &
        Hybrid
        &
        81.2
        &
        80.9
        &
        80.3
        \\
        \rowcolor{blue!12}&
        \textbf{Merging} %
        &
        \textbf{81.2}
        &
        \textbf{81.0}
        &
        \textbf{80.3}
        \\
    \bottomrule
    \end{tabular}
    \caption{
    Ablation study of different reduction choices on top-1 accuracy (\%). Pruning involves directly removing tokens from the 'Source' group, while Merging refers to our method of combining 'Source' tokens with those in the 'Target' group. Hybrid combines both Pruning and Merging methods; for simplicity, we allocate 50\% of the tokens to each strategy.
    }
    \label{tab:exp_ab_operation}
\end{table}

\noindent \textbf{Analysis on reduction operation.}
Unlike previous approaches that treat the hidden state and residual in Mamba separately~\cite{zhan2024rethinking, zhan2024exploring}, our approach applies the same reduction strategy to both the hidden state and residual, ensuring simplicity and information consistency. Table~\ref{tab:exp_ab_operation} presents experiments on different reduction strategies. The results indicate that the merging strategy outperformed other reduction methods, as it minimizes information loss. Additionally, our results indicate that with stronger models or smaller reduction ratios, even the pruning strategy does not significantly impact performance, confirming that the filtered 'Source' tokens are indeed unimportant. \textit{More experiments on reduction strategies are provided in the Appendix.}

\begin{figure*}[t!]
    \centering
    \includegraphics[width=0.90\linewidth]{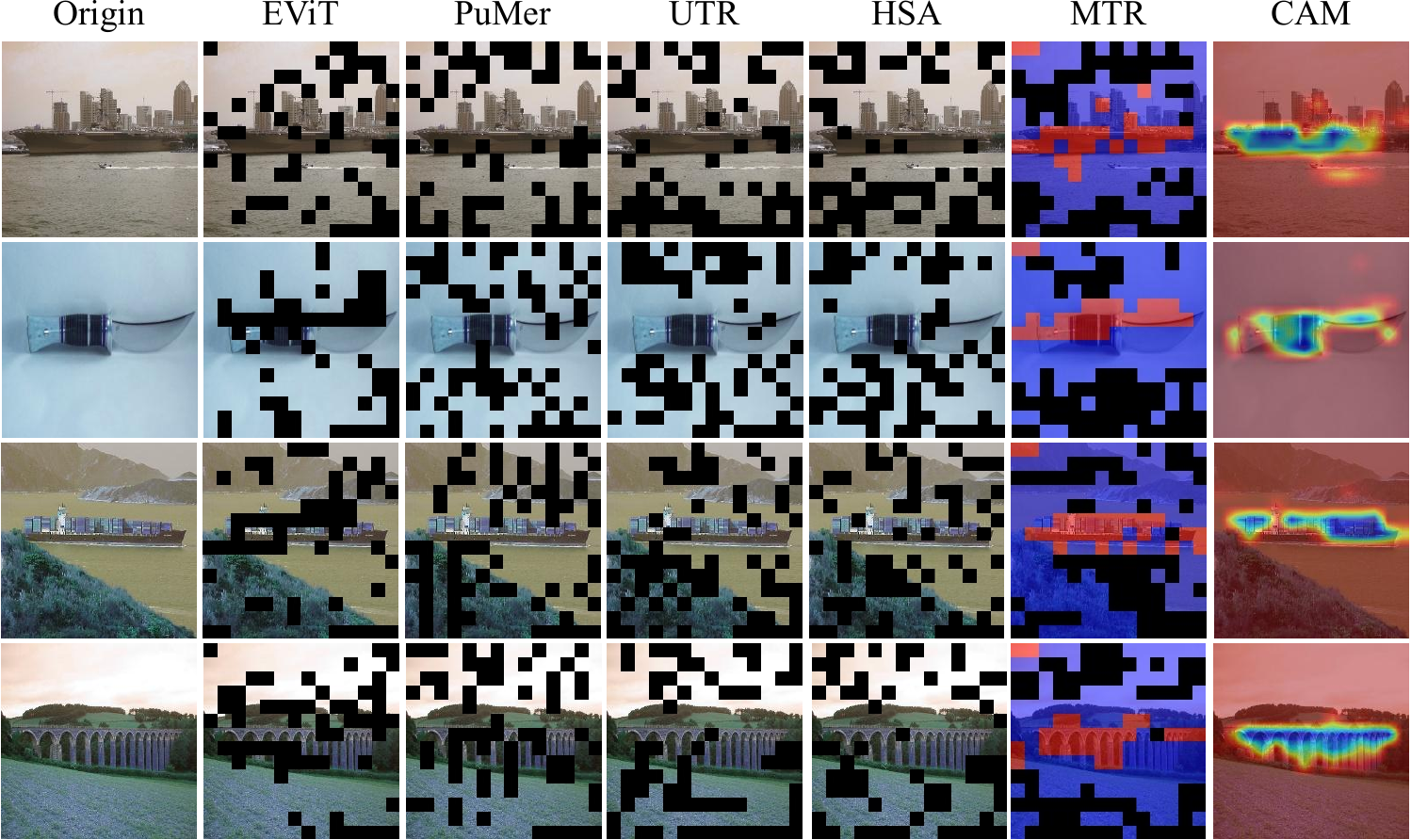}
    \caption{Visualization of reduction tokens on ViM-S under 20\% overall reduction of FLOPS. We present visualizations of the original image and the corresponding image after token reduction for each method. The masked regions represent the reduction tokens. For our MTR, the red tokens indicate those in the 'Keep' group, while the blue tokens indicate the 'Target' group, and the masked tokens represent the 'Source' group. We also display Class Activation Maps (CAM) in the rightmost column. Notably, the yellow and blue areas in the CAM diagram indicate highly responsive regions, while the red areas indicate low responsive regions.
    }
    \label{fig_3}
\end{figure*}

\begin{figure}[!t]
  \centering
  \includegraphics[width=0.45\textwidth]{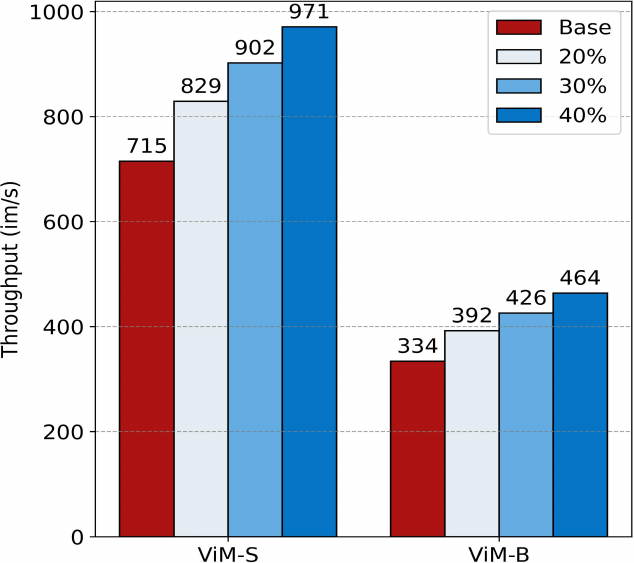}
  % includesvg
  % \includesvg[width=3.0in]{figures/figure1.svg}
  \caption{Comparison of generation throughput across different FLOPS reduction ratios for ViM-S and ViM-B.}
\label{fig_4}
\vspace{-0.2cm}
% \end{wrapfigure}
\end{figure}

\noindent \textbf{Visualization.} To further investigate the interpretability of MTR, we visualize the retained visual tokens in various scenarios in Fig.~\ref{fig_3}. We present the original images and the retained visual tokens of different methods. It can be observed that the red tokens in MTR essentially correspond to the most responsive regions in the CAM. This indicates that the tokens within our ‘Keep’ group align with the image's core content. Moreover, foreground objects are primarily encompassed within red or blue tokens, while black tokens predominantly represent task-irrelevant regions. This suggests that MTR effectively retains category-specific tokens and excludes irrelevant background tokens. It is worth noting that UTR and HSA, which are also importance-based token reduction methods, still exclude some tokens related to the foreground.

\noindent \textbf{Inference throughput.} As our approach compresses the input token number, we can accelerate inference and achieve higher model throughput, as illustrated in Fig.~\ref{fig_4}. The throughput increases with the reduction ratio. By adjusting the reduction ratio, we can choose to prioritize model performance, inference speed, or a balance of both.

\section{Conclusion}
In conclusion, this paper introduces a training-free Mamba token reduction framework, MTR, addressing the incompatibility of existing Vision Transformer (ViT) token reduction methods with Mamba, which lacks attention mechanisms and relies on token order. To solve these challenges, MTR leverages Mamba’s internal timescale parameter $\Delta$ to assess token importance, groups tokens by importance into 'Keep,' 'Target,' and 'Source' categories, and merges similar tokens while preserving order. The proposed MTR framework can be easily adapted to Mamba models without introducing additional parameters or requiring a training process. Extensive experiments demonstrate that MTR achieves state-of-the-art performance across various benchmarks and significant inference acceleration, underscoring its superiority.

\bibliography{MTR}

\end{document}